\documentclass{article}

\usepackage{arxiv}

\usepackage[utf8]{inputenc} 
\usepackage[T1]{fontenc}    
\usepackage{hyperref}       
\usepackage{url}            
\usepackage{amsmath}
\usepackage{booktabs}       
\usepackage{amsfonts}       
\usepackage{nicefrac}       
\usepackage{microtype}      
\usepackage{lipsum}		
\usepackage{graphicx}
\usepackage{times}
\usepackage{epsfig}
\usepackage{graphicx}
\usepackage{amsmath}
\usepackage{amssymb}
\title{Real-Time Object Detection and Localization in Compressive Sensed Video on Embedded Hardware}
\newcommand{\etal}{\textit{et al}.}

\author{
  Yeshwanth Ravi Theja Bethi \\
  Department of Electronics and Systems Engineering \\
  Indian Institue of Science \\
  Bangalore, India 560012 \\
  \texttt{yeshwanthb@iisc.ac.in} \\
   \And
 Sathyaprakash Narayanan \\
Department of Electronics and Systems Engineering \\
  Indian Institue of Science \\
  Bangalore, India 560012 \\
  \texttt{sathyaprakas@iisc.ac.in} \\
  \AND
  Venkat Rangan \\
 tinyVision.ai Inc. \\
  San Diego, California \\
  Bangalore, India 560012 \\
  \texttt{venkarrangan2005@gmail.com} \\
  \And
  Chetan Singh Thakur \\
  Assistant Professor \\
  Department of Electronics and Systems Engineering \\
  Indian Institue of Science \\
  Bangalore, India 560012 \\
  \texttt{csthakur@iisc.ac.in}
}
\begin{document}
\maketitle

\begin{abstract}
Everyday around the world, interminable terabytes of data are being captured for surveillance purposes. A typical 1-2MP CCTV camera generates around 7-12GB of data per day. Frame-by-frame processing of such enormous amount of data requires hefty computational resources. In recent years, compressive sensing approaches have shown impressive results in signal processing by reducing the sampling bandwidth. Different sampling mechanisms were developed to incorporate compressive sensing in image and video acquisition. Pixel-wise coded exposure is one among the promising sensing paradigms for capturing videos in the compressed domain, which was also realized into an all-CMOS sensor \cite{Xiong2017}. Though cameras that perform compressive sensing save a lot of bandwidth at the time of sampling and minimize the memory required to store videos, we cannot do much in terms of processing until the videos are reconstructed to the original frames. But, reconstruction of compressive-sensed (CS) videos still takes a lot of time and is also computationally expensive. In this work, we show that object detection and localization can be possible directly on the CS frames (easily upto 20x compression). To our knowledge, this is the first time that the problem of object detection and localization on CS frames has been attempted. Hence, we also created a dataset for training in the CS domain. We were able to achieve a good accuracy of 46.27\% mAP(Mean Average Precision) with the proposed model with an inference time of 23ms directly on the compressed frames(approx. 20 original domain frames), this facilitated for real-time inference which was verified on NVIDIA TX2 embedded board. Our framework will significantly reduce the communication bandwidth, and thus reduction in power as the video compression will be done at the image sensor processing core.
\end{abstract}

\section{Introduction}
\label{sec:intro}
{\indent} In signal processing, compressive sensing \cite{Baraniuk2017} is a powerful sensing paradigm to sample sparse signals with much fewer samples than demanded by the Shannon-Nyquist sampling theorem. The Nyquist theorem mandates that the number of data samples to be at least as big as the dimensionality of the signal being sampled. Inherent redundancy present in the real  signals like images and videos allows significant compression of data. Compressive sensing exploits this inherent redundancy and enables the sampling to happen at sub-Nyquist rates. As generic video acquisition systems face a trade-off between spatial and temporal resolution due to communication bandwidth, systems that can give higher spatial resolution without compromising temporal resolution are in high demand.

\begin{figure*}[!ht]
\begin{center}
\fbox{\includegraphics[width=\textwidth]{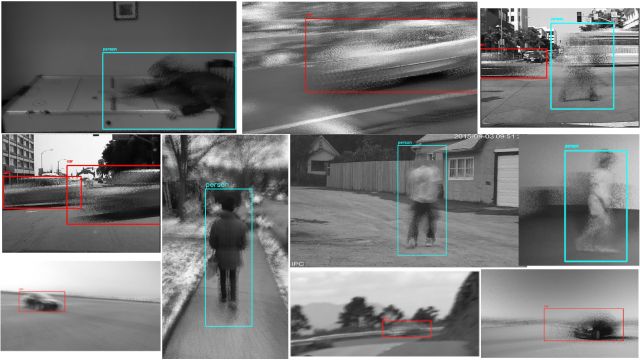}}
\caption{Object detection and localisation in CS frames by bounding boxes. Cyan boxes show persons, and red boxes show cars.}
\label{fig:teaser}
\end{center}

\end{figure*}

This makes compressive sensing extremely useful for capturing images and videos in systems that cannot afford high data bandwidth as the number of samples needed for the same duration of video is much lower than that of generic imaging systems. The ability to sample very few data points and still be able to reconstruct the original signal helps to create low-power imaging systems. Pixel-wise coding \cite{Hitomi2011} is one among many ways of compressive sampling the videos. Each pixel is independently exposed for a specific amount of time ~$T_b$ (bump time) and the start time of that exposure is decided by a random sensing matrix.

Using this controlled exposure, the imaging system can obtain a slower frame rate without having to worry about motion blur. Typically a 100 FPS video can be reconstructed from a 5 FPS video, using pixel-wise coded exposure. Jie Zhang \etal \cite{Zhang2016} have created an all-CMOS spatio-temporal CS video camera with pixel-wise coded exposure technique at its core.

To reconstruct the original frames of a video from CS frames, one needs to learn an over-complete dictionary or use existing dictionaries such as 3D DCT dictionary to represent the videos in a sparse domain. The reconstruction of each CS frame is done on a patch-by-patch basis. A patch of size around ~$7\times7$ is taken from the CS frame and the corresponding video patch in the sparse domain of the dictionaries is estimated using algorithms like Orthogonal Matching Pursuit (OMP). Recent works such as Iliadis \etal \cite{Iliadis2016} and Xu \etal \cite{Xu2016} have even succeeded to reconstruct original frames from CS frames using deep learning.
Though this line of work is still in progress, reconstruction of the CS frames takes considerable amount of time, which makes it unsuitable for any sort of real-time processing. To overcome this issue, we propose a new approach by eliminating the need to reconstruct the original frames for object detection and localization by performing it directly on the compressed frame.
We simulated the images created by a CS camera \cite{Xiong2017} \cite{Zhang2016} by compressing 30 FPS videos with 13x compression using a randomly generated sensing matrix with a bump time of 3 frames. We generated a dataset for the object detection task using the camera simulator.

\section{Methods}
\label{sec:methods}
\subsection{Pixel-wise coded exposure}
In conventional cameras, there is a global exposure time ~$T_{e}$ (shutter speed) for which all the pixels in the sensor are exposed and an image is readout from the sensor. We get ~$1/T_{e}$ frame rate from such sensors. In contrast to this, in pixel-wise coded exposure (PCE) cameras, each pixel is exposed at a random time for a 'single-on' fixed duration ~$T_{b}$ (> ~$T_{e}$) within the time ~$T_{v}$. But, only one image is read out at the end of ~$T_{v}$. If ~$T_{v}$ is set to ~$C \times T_{e}$, we get a compression of ~$C$ times and a frame rate of ~$1/T_{v}$. This single frame at the end of ~$T_{v}$ has all the necessary information of the intermediate frames encoded in it.    

\begin{figure*}[!ht]
\begin{center}
\fbox{\includegraphics[width=\textwidth]{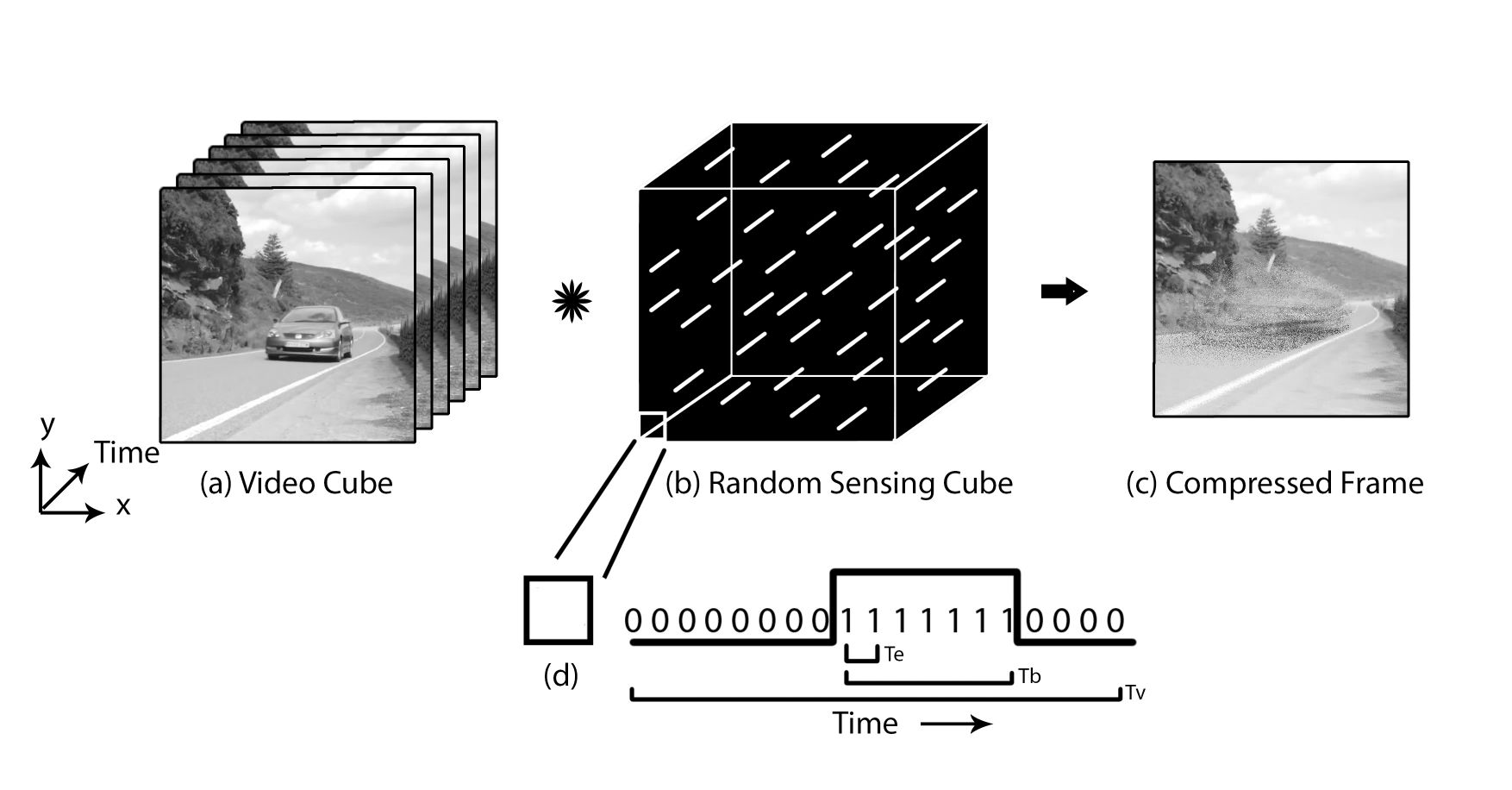}}
\caption{Pixel-wise coded exposure}
\label{fig:pce}
\end{center}
\end{figure*}

Figure \ref{fig:pce} shows the process of PCE. Though this creates only ~$1/T_{v}$ frame rate, all the intermediate frames (~$1/T_{e}$) can be reconstructed using the OMP algorithm based on the random sensing times of each pixel (Sensing Matrix). This makes it suitable for capturing high temporal resolution video without compromising the spatial resolution.

\subsection{Sensing Matrix}
Consider a spatio-temporal video ~$V ( M \times N \times T )$. A sensing matrix ~$S ( M \times N \times T )$ holds the exposure control values for each pixel in the sensor. The sensing matrix ~$S$ is binary, i.e., ~$S \in \{1,0\}$. The value of ~$S$ is 1 for each pixel for the frames for which it is on or exposed, and 0 for the rest of the frames. There is only one bump in the ~$T_{v}$ time. 

The position of 1s in S is randomly decided based on a Gaussian normal distribution, i.e, the start time of each pixel's exposure time is randomly chosen for each pixel and then it is exposed for ~$T_{b}$ (Bump Time) number of frames.
These values are then summed up along the time dimension. The acquired coded image ~$I ( M \times N)$ can be denoted by the equation:
\begin{equation}
\label{eq:cs}
I (m,n) = \sum_{t=1}^{T} S(m,n,t).V(m,n,t)
\end{equation}

\subsection{Dataset}

\begin{figure}[!ht]
\centering
\fbox{\includegraphics[width=8cm,height=5cm,keepaspectratio]{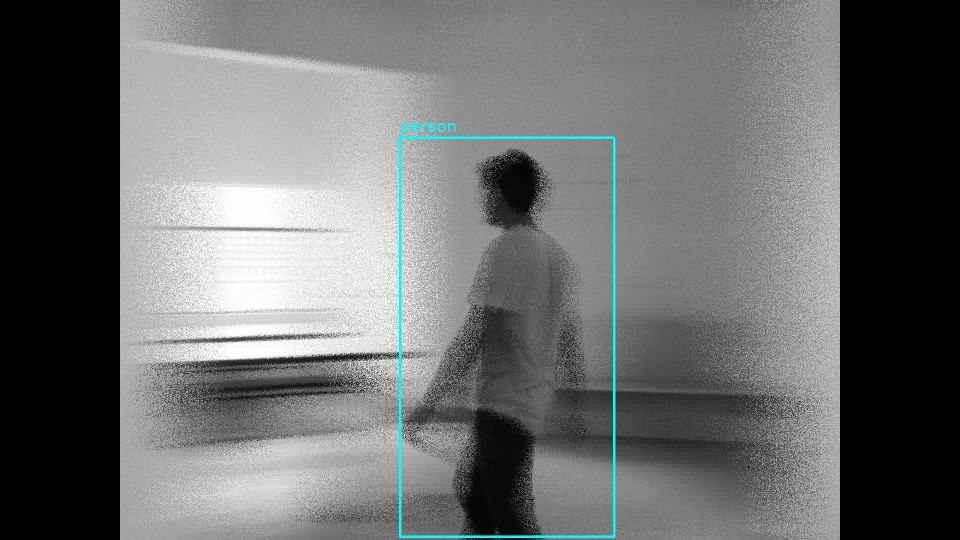}}
\caption{Labelled compressed frame }
\label{fig:cs_d}
\end{figure}

Due to the unavailability of any pre-existing public dataset on temporally compressed videos, we had to create our own dataset. We searched the Youtube 8M \cite{Youtube8M} dataset for videos which had either a single person or a single car present in it. As all the variations were not available, we collected more videos on our own within our university campus for more training data. We made sure that only one object of interest was present in all of the videos captured. We captured videos on multiple days and during different parts of the day and tried to maintain as much variance as possible. The videos include moving objects with a stationary background, stationary objects with moving background and both the object and background moving(Figure \ref{fig:cs_d}).
We also considered the case where the frame (compressive sensing camera) used to capture is also in motion with respect to the scene. This dataset comprises a combination of motion, both with respect to the camera and the object, thus covering the entire spectrum of instances one can witness in the original domain to be addressed in the compressed domain.

\begin{enumerate}
 \item \textbf{Methodology} \\
 \begin{enumerate}
  \item \textbf{Compression Technique using PCE}   \\
  
  \begin{figure*}[!ht]
\begin{center}
\fbox{\includegraphics[scale=0.5]{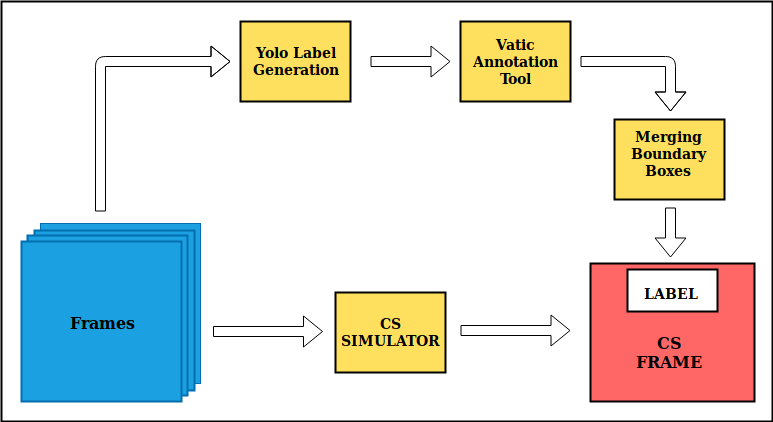}}
\caption{Pipeline for compression}
\label{fig:pipeline}
\end{center}

\end{figure*}

30 FPS clips were accumulated from various sources and compressed with 13X compression rate. The bump time ~$T_{b}$ is set to 3. For every 13 frames in the original video, a single compressed frame is generated using equation\ref{eq:cs}. The random sensing matrix has been varied for each set of 13 frames, so as to generalize compressed sensed frames and to not fit over a particular sensing matrix. We normalize the pixel values after adding them through 13 frames to constrain the pixel value within 255. We only worked with monochrome videos because conventionally only monochromatic CMOS sensors \cite{Xiong2017} are available.

\item \textbf{Psuedo labels for ground truth} \\
Manually labeling of the CS frames is a difficult job as the edges of moving objects are often indeterminable even to human eye. To avoid this problem, we chose pseudo-labeling of the data. As most of the existing architectures like YOLO v3\cite{redmon2018yolov3} detect persons and cars classes with really good accuracy in the original domain, we used trained YOLO v3 to pseudo label the CS frames Fig.\ref{fig:cs_d}. 

\begin{figure*}
\begin{center}
\fbox{\includegraphics[width=\textwidth]{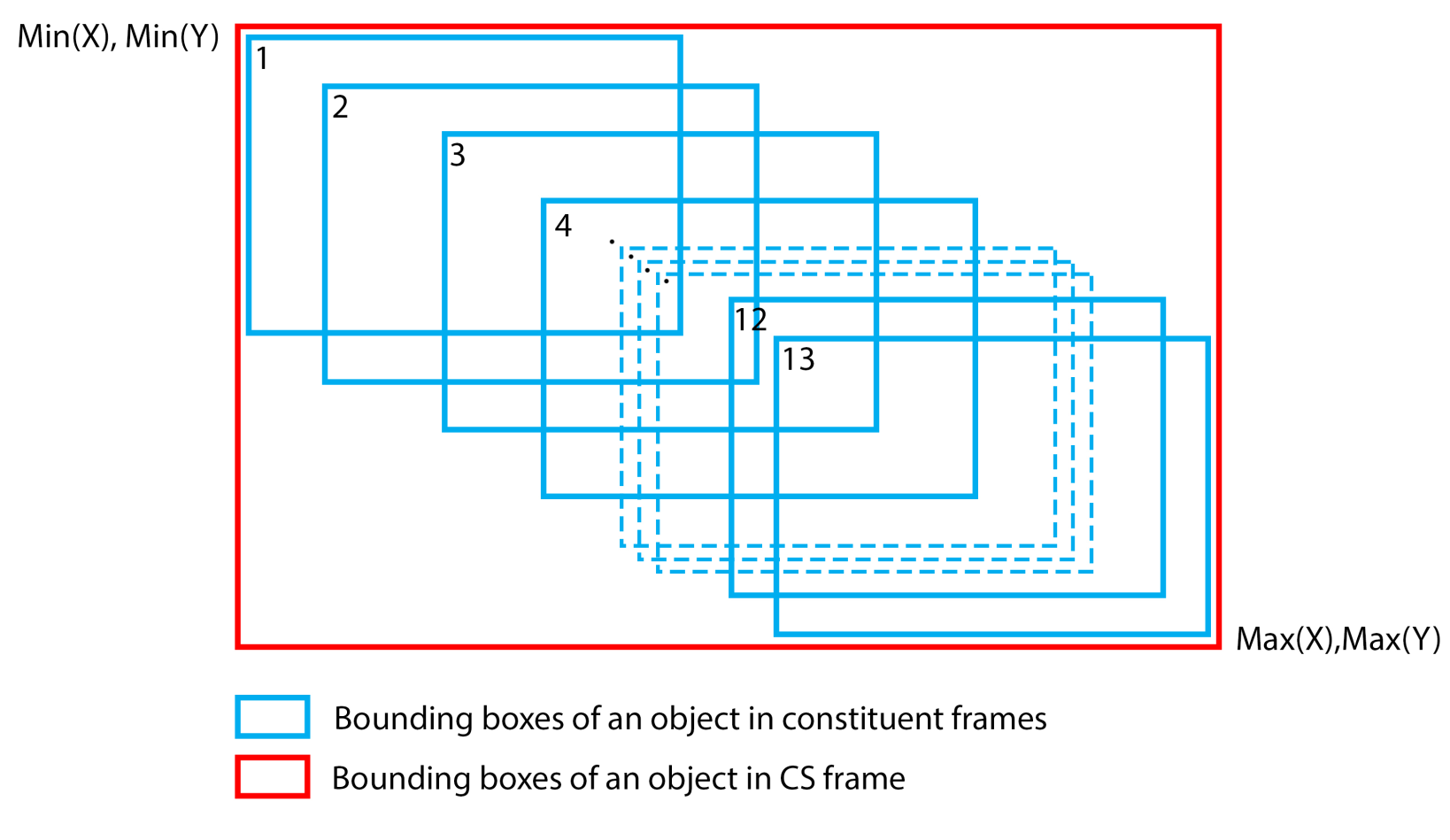}}
\caption{Pseudo labelling}
\label{fig:pseudo_labelling}
\end{center}

\end{figure*}

\begin{figure*}[!ht]
\begin{center}
\fbox{\includegraphics[width=\textwidth]{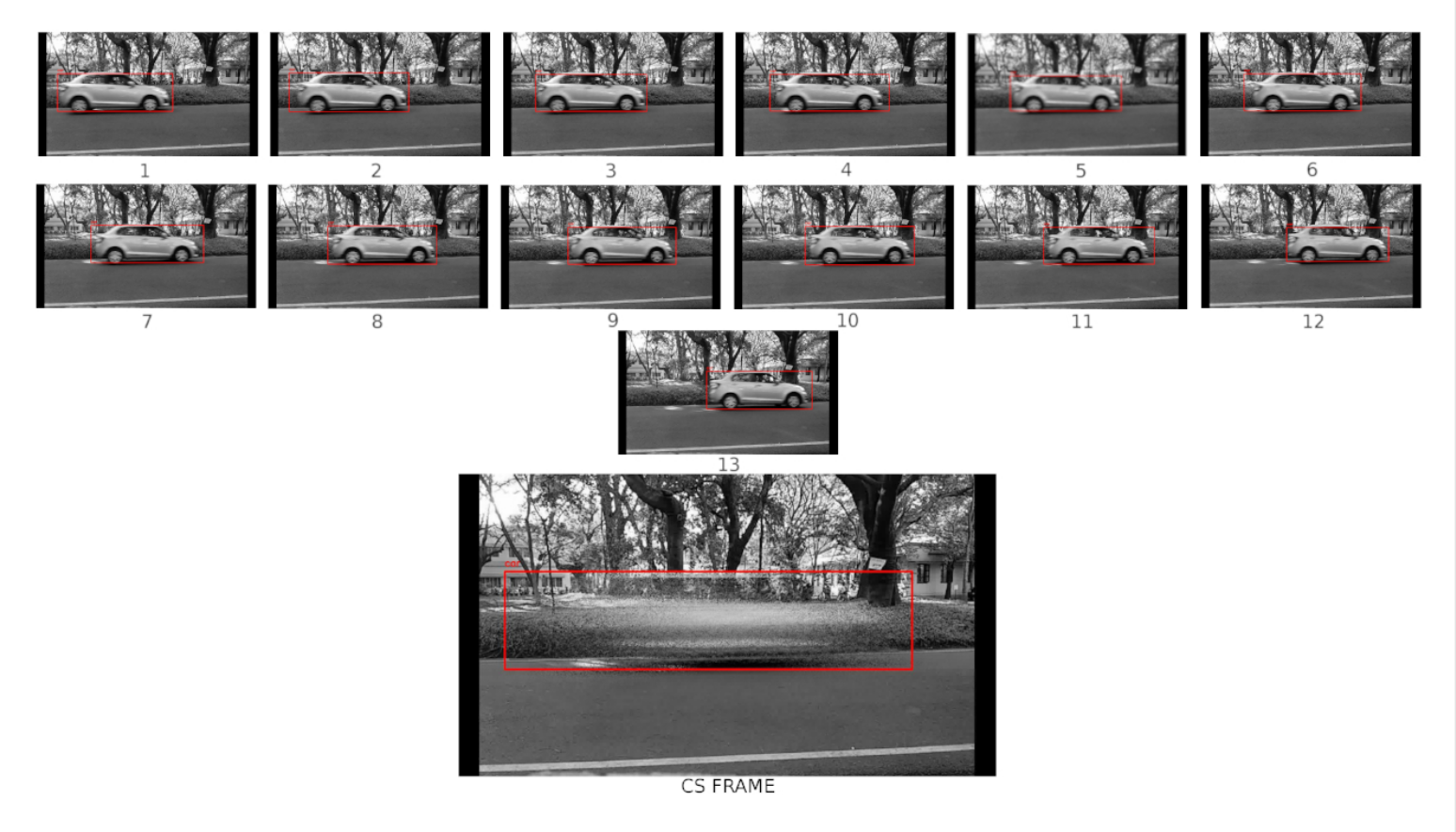}}
\caption{Merging boundary box}
\label{fig:merging_bbox}
\end{center}

\end{figure*}

As shown in Fig.\ref{fig:pipeline} we used a pre-trained YOLO model to generate the bounding boxes of the individual frames. The threshold for detection was kept more than 99\%, so as avoid any ambiguous predictions. We fed these pseudo labeled boxes generated by YOLO with the corresponding frames onto VATIC\cite{vondrick2013efficiently} annotation module, where it was manually corrected for the boundary boxes to fit it more compactly. Manual labeling was considered to mitigate the delta difference possessed with the YOLO prediction from the actual ground truth.

We also had to come up with a way to draw bounding boxes in the CS domain. So, we chose to draw a bigger bounding box in the CS frame which enclosed all the individual bounding boxes of the constituent frames. Fig.\ref{fig:merging_bbox} shows the method to merge the bounding boxes of each object across the constituent frames. Then with these boundary box coordinates, we find the minimum and maximum of its X and Y coordinates for the entire clip's timeline. These coordinates are then used for the bigger bounding box in the CS frame Fig.\ref{fig:pseudo_labelling}. Tracking a single object's detection in consecutive frames which contain multiple objects of the same class is difficult because YOLO doesn't order them in any particular fashion. Because of this difficulty to attach each bounding box and track it along the frames to a specific object, for training dataset we collected videos which have only one object of interest. This kind of labeling gave a preemptive measure to easily rectify and provide a tight fit of boundary box for the object in the frame with ease. Thus allowing us to create a large dataset of annotated compressive sensed images, these labelled boundary boxes were then combined into larger bounding boxes that were used as the ground truth values of the bounding boxes in the CS frames Fig.\ref{fig:merging_bbox}. 
We wanted less human intervention for the labeling, the pseudo labeling provided an extra hand for the human introspection for labeling. By providing the pseudo labeling the human effort here was reduced and more attention was drawn towards reiterating the boundary box more tightly to fit the objects more precisely on the compressed frame Fig.\ref{fig:merging_bbox}. Approximately 45,472 and 60,708 datapoints samples are available for the class Cars and Person respectively. This dataset is now openly available for the community.

 \end{enumerate}
\end{enumerate}

\subsection{Model and Training}

\begin{figure}[!ht]
\centering
\fbox{\includegraphics[width=19cm,height=6cm,keepaspectratio]{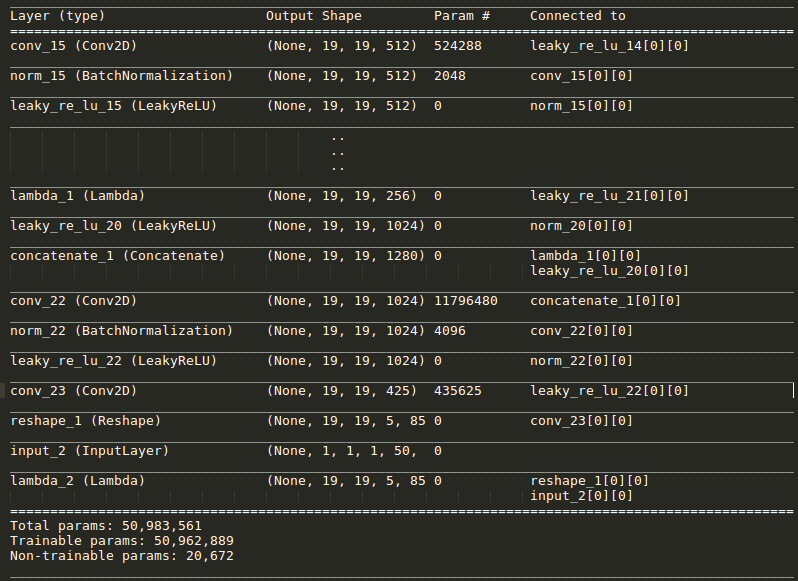}}
\caption{Layers used for detection in the Proposed Model}
\label{fig:cs_model}
\end{figure}

We used the pre-trained initial layers of YOLO(14 layers), these layers pertain the prior knowledge for the detection and localization of objects in the original RGB domain. We passed the output of these layers through a series of 7 Convolution layers(resembling Layer 15) as described in the Fig. \ref{fig:cs_model}, followed by BatchNormalization \cite{ioffe2015batch} and composed them onto a 256 dimensional vector which was activated by LeakyReLU \cite{xu2015empirical}. Now, the output of the activation layer(LeakyReLU) and the lambda layer are concatenated, this is performed to ensure that the model can comprehend each instance of the object present from the original domain and triangulate the location and the class label of the object present in the compressed domain. This latent space is then fed to a Convolution layer that deciphers the presence of the object and location of the object and the class labels present in the compressed frame. The cars or persons in the CS frames have high variance in their appearance, based on the level of motion present in the 13 frames. Depending on the motion, they can either look like normal cars/persons when stationary or hazy when in motion. Therefore, we needed a pre-trained model with prior knowledge similar to that of YOLO v2 to differentiate the different classes (Cars and Persons being one among them). The model being trained from scratch without pre-trained weights, has the risk of fitting to the blurred patterns of cars and persons. Hence, we generated large training dataset and also started the training with the pre-trained weights.

The model was trained using Tensorflow\cite{Abadi2016} and Keras \cite{chollet2015keras}. The training converges within 25 epochs, using Adam optimizer \cite{Kingma2014} for parameter updates with learning rate = 1e-5, beta\_1 = 0.9, beta\_2 = 0.999, and epsilon = 1e-08. The mini-batch size was 16 for the training. We trained this hybrid network on the dataset with a train and test split ratio of  7:3.

\section{Results and Discussion}
\label{sec:results}
\subsection{Initial Results}
Initially, we attempted to train only for a single class of objects. The model gave very poor results in terms of output confidence levels and IOUs. It was not able to differentiate between the moving background and the moving object. However, with both the classes in the training set, the model gave much better results and high output activations when compared to models trained for a single class. This proves that adding more classes increases the capability of the model to learn the differentiating features. Final results of the model on various dynamic scenes of both classes can be seen in Figures \ref{fig:cars} \& \ref{fig:persons}.

\subsection{Mean Average Precision}
We have followed the COCO dataset's\cite{COCO} method of evaluating the Mean Average Precision (mAP), by averaging the Average Precisions (APs) over multiple IOU (Intersection over Union ratio of predicted box and ground truth) thresholds from 0.50 to 0.95. The 'Car' and 'Person' classes got an mAP of 42.42\% and 50.12\%, respectively. We also experimented with our trained model to test on CS frames with different compression rates and bump times. We chose sequences of CS frames that gave a good detection with 13x compression (at which the model was trained) and recreated the same sequences with different compression and bump time rates.
The AP values at different IOUs for different variations of bump times and compression rates are shown in Table \ref{tab:bump_time} and \ref{tab:cmp_rates}, respectively.

\begin{table}[!ht]
\begin{center}
\begin{tabular}{|l|c|}
\hline
Model & Mean Average Precision (mAP) \\
\hline\hline
Proposed Model & 46.27 \\
YOLO v3 & 7.53 \\
SNIPER &  7.61\\
\hline
\end{tabular}
\end{center}
\caption{mAP comparison with custom dataset on few state of the art models and the proposed model}
\label{tab:sota}
\end{table}

The results showed no conclusive proof of any trend of decline or ascent in the mAP with respect to the bump time or the compression unless they are the extremes. This shows that our model is generalized and can work with other compression rates and bump times.

\begin{table*}[!ht]
\begin{center}
\scalebox{0.51}[1]{
\begin{tabular}{|c|c|c|c|c|c|c|c|c|c|c|c|}
\hline
Bump Time & AP at IOU = 0.50 & AP at IOU = 0.55 & AP at IOU = 0.60 & AP at IOU = 0.65 & AP at IOU = 0.70 & AP at IOU = 0.75 & AP at IOU = 0.80 & AP at IOU = 0.85 & AP at IOU = 0.90 & AP at IOU = 0.95 & Mean AP \\ 
\hline\hline
2 & 1 &	1 &	1 &	0.95 & 0.95 & 0.93 & 0.81 & 0.55 & 0.40 & 0.03 & 0.76
 \\
3 &	1 &	1 &	1 &	0.93 & 0.93 & 0.91 & 0.80 & 0.50 & 0.37 & 0 & 0.74
 \\
4 &	1 &	1 &	1 &	0.94 & 0.94 & 0.92 & 0.81 &	0.5 & 0.25 & 0.01 &	0.74
\\
5 &	1 &	1 &	1 &	0.95 & 0.95 & 0.93 & 0.82 &	0.51 & 0.29 & 0.02 & 0.75
\\
\hline

\end{tabular}}
\caption{Average precisions (AP) at different IOU and mAP with change in Bump times Tb.}
\label{tab:bump_time}
\end{center}
\end{table*}

\begin{table*}[!ht]
\begin{center}
\scalebox{0.50}[1]{
\begin{tabular}{|c|c|c|c|c|c|c|c|c|c|c|c|}
\hline
Compression Rate & AP at IOU = 0.50 & AP at IOU = 0.55 & AP at IOU = 0.60 & AP at IOU = 0.65 & AP at IOU = 0.70 & AP at IOU = 0.75 & AP at IOU = 0.80 & AP at IOU = 0.85 & AP at IOU = 0.90 & AP at IOU = 0.95 & Mean AP\\
\hline\hline
6 &	1 &	1 &	1 &	1 &	1 &	1 &	0.33 & 0.08 & 0.08 &	0 &	0.64
 \\
10 & 1 & 1 & 1 & 1 & 1 & 1 & 0.8 & 0.35 & 0.2 &	0.02 &	0.74
 \\
13 & 1 & 1 & 1 & 0.93 & 0.93 & 0.91 & 0.80 &	0.50 & 0.37 & 0 & 0.74
\\
16 & 1 & 1 & 1 & 0.90 & 0.90 & 0.90 & 0.86 & 0.55 & 0.36 & 0 & 0.75
\\
20 & 1 & 1 & 1 & 0.88 & 0.87 & 0.85 & 0.84 & 0.52 & 0.32 & 0.01 &	0.73
\\
24 & 1 & 1 & 0.97 & 0.89 & 0.83 &	0.83 & 0.82 & 0.57 & 0.30 &	0.03 & 0.72
\\
\hline

\end{tabular}}
\end{center}
\caption{Average precisions (AP) at different IOU and mAP with change in compression rates.}
\label{tab:cmp_rates}
\end{table*}

\subsection{Discussion}

\begin{figure*}[!ht]
\begin{center}
\begin{tabular}{ccc}
\fbox{\includegraphics[width=3.5cm]{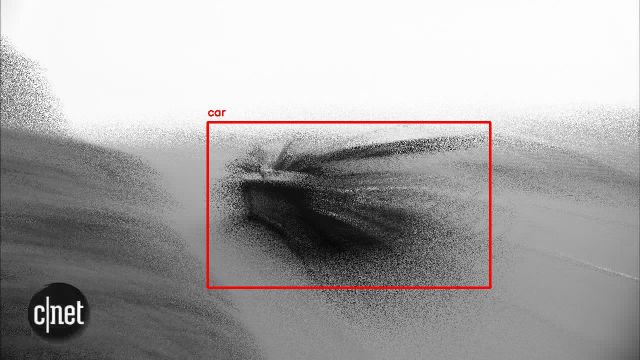}}&
\fbox{\includegraphics[width=3.5cm]{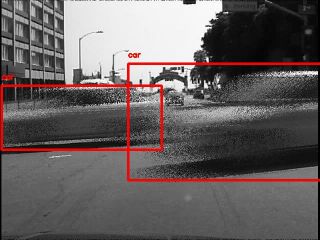}}&
\fbox{\includegraphics[width=3.5cm]{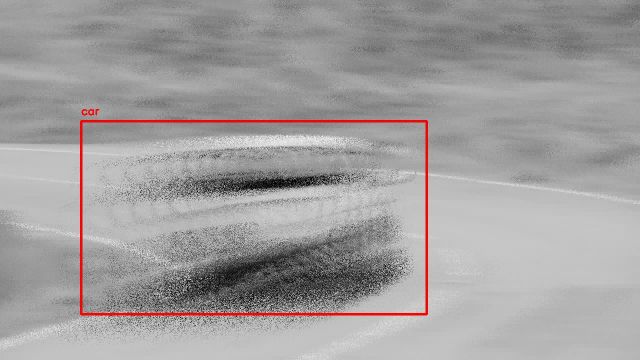}}\\
(a)&(b)&(c)\\
\fbox{\includegraphics[width=3.5cm]{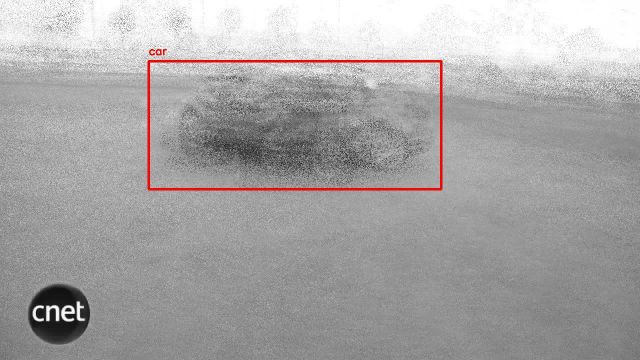}}&
\fbox{\includegraphics[width=3.5cm]{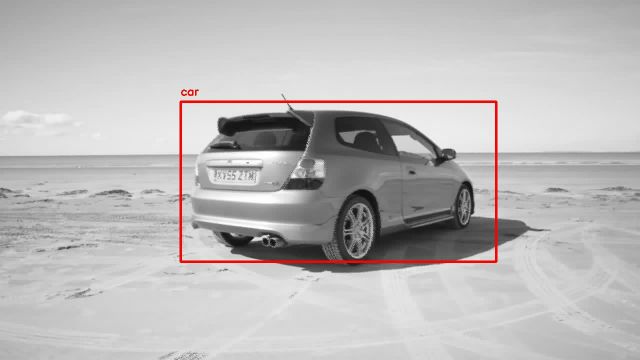}}&
\fbox{\includegraphics[width=3.5cm]{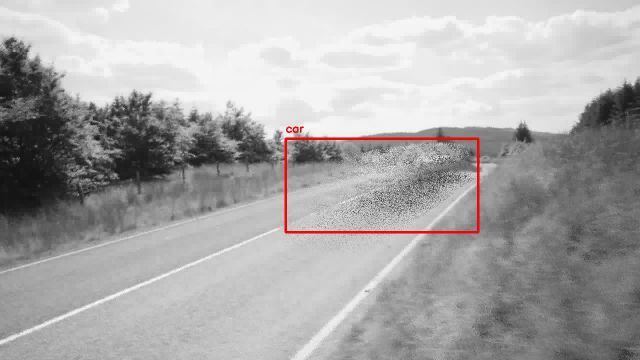}}\\
(d)&(e)&(f)
\end{tabular}
\end{center}
\caption{Results on cars. (a), (c), (d) Car and background both in motion; (b), (f) Car in motion with a stationary background; (e) Still car and background.}
\label{fig:cars}
\end{figure*}

\begin{figure*}[!ht]
\begin{center}
\begin{tabular}{ccc}
\fbox{\includegraphics[width=3.5cm]{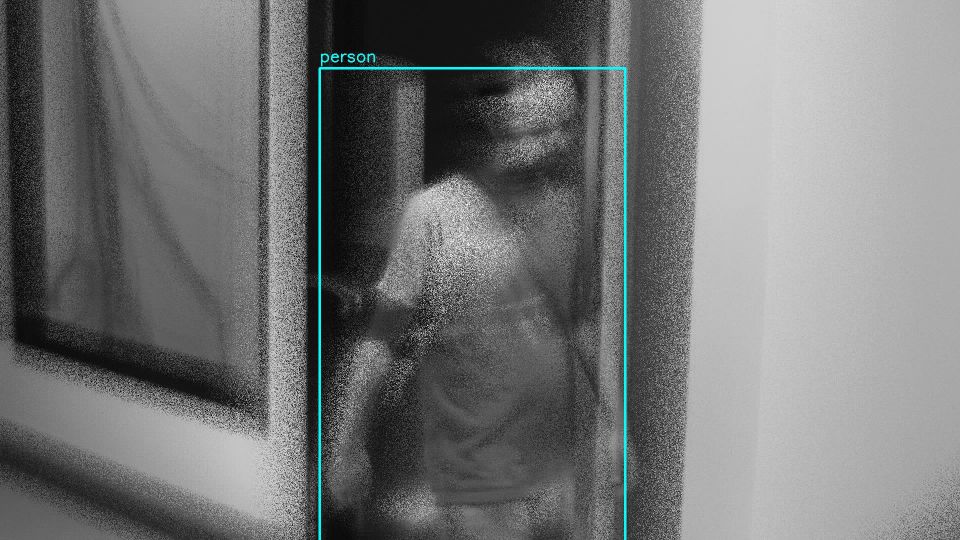}}&
\fbox{\includegraphics[width=3.5cm]{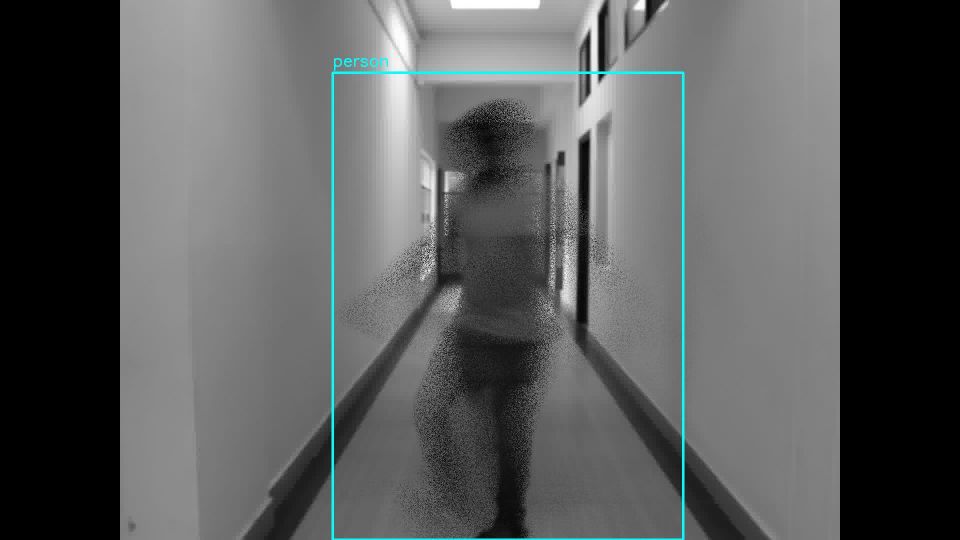}}&
\fbox{\includegraphics[width=3.5cm]{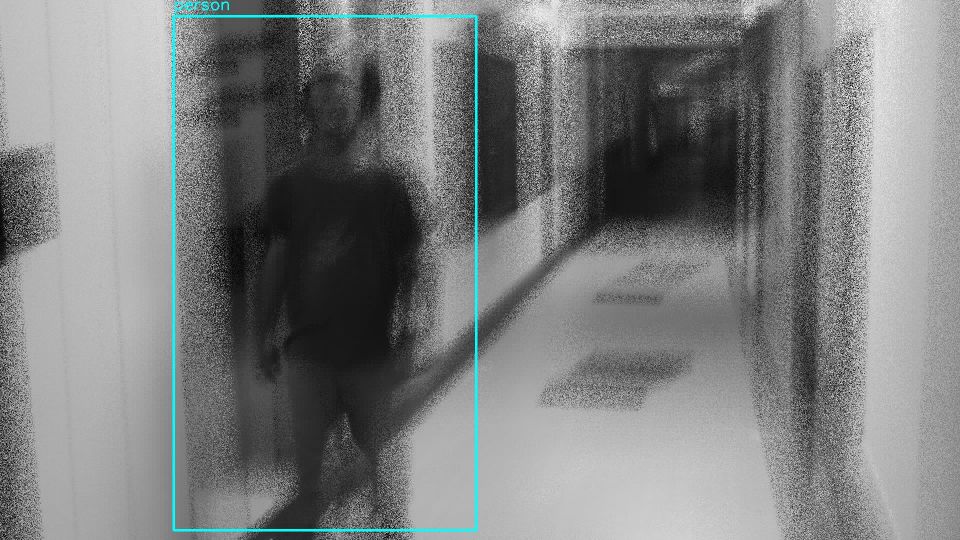}}\\
(a)&(b)&(c)\\
\fbox{\includegraphics[width=3.5cm]{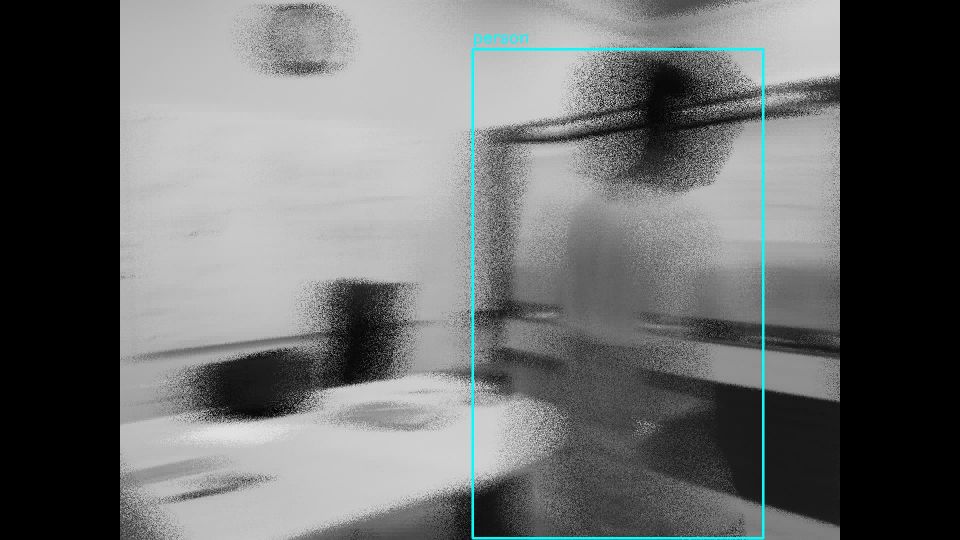}}&
\fbox{\includegraphics[width=3.5cm]{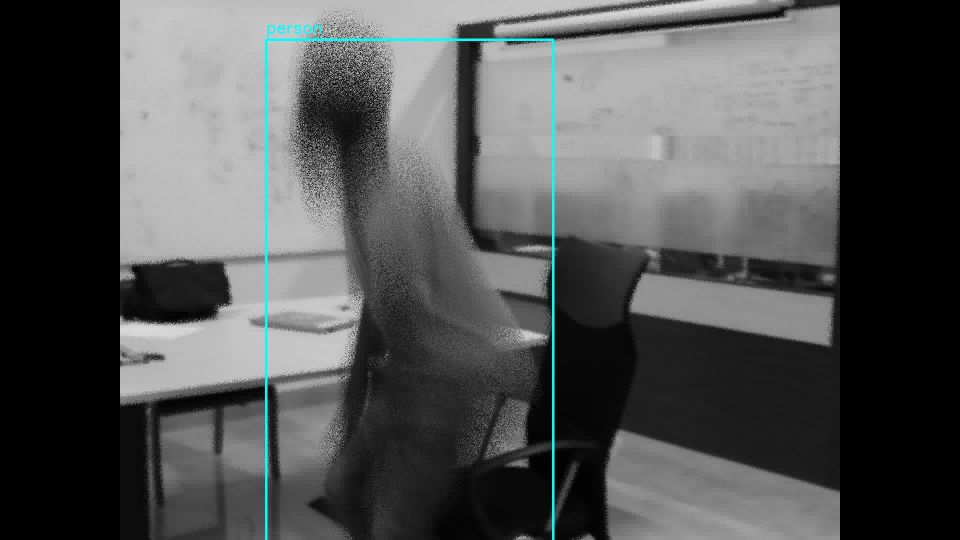}}&
\fbox{\includegraphics[width=3.5cm]{person6.jpg}}\\
(d)&(e)&(f)
\end{tabular}
\caption{Results on persons. (a), (c), (d) Person and background both in motion; (b), (e) Person moving, with a stationary background; (f) Almost still person with moving background.}
\label{fig:persons}
\end{center}
\end{figure*}

\begin{figure*}[!ht]
\begin{center}
\begin{tabular}{ccc}
\fbox{\includegraphics[width=3.5cm]{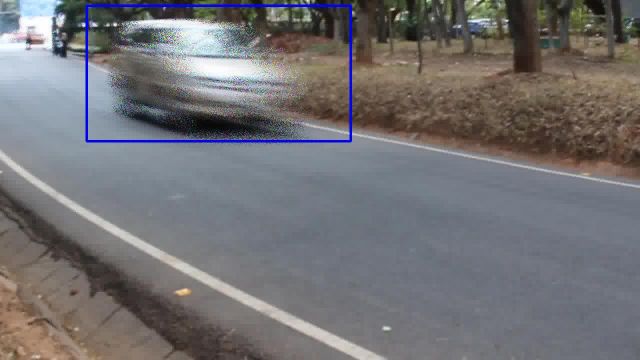}}&
\fbox{\includegraphics[width=3.5cm]{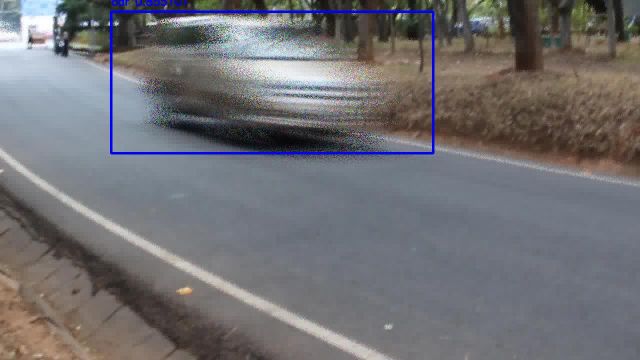}}&
\fbox{\includegraphics[width=3.5cm]{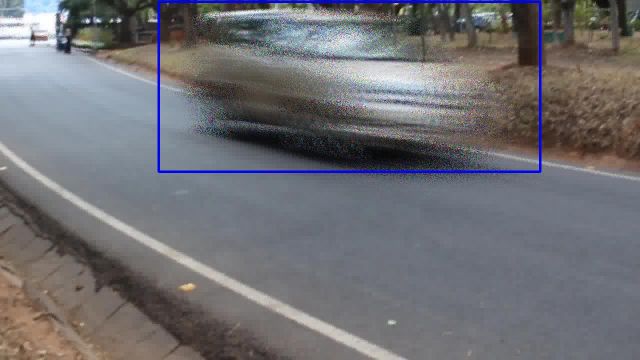}}\\
\end{tabular}
\caption{Detection on colour CS frames.}
\label{fig:colour}
\end{center}
\end{figure*}

We can observe from the Tables \ref{tab:cmp_rates} and \ref{tab:bump_time}, that different compression rates of the CS frames do not change the mAP drastically. The reason why other compression rates also work is because, CS frames have high variance in the appearance of the objects, based on the amount of motion in the constituent frames. If the object is stationary and still all through the frames, the object in the corresponding CS frame also appears as seen in Figure~\ref{fig:cars}(e). In such cases, irrespective of the compression or the bump time rates, the CS frames look very alike. The same can happen with combinations of different amounts of motion at different compression rates. Hence, once a model is trained on varying levels of motion at a compression rate, it can also work with other compression rates fairly well. We also compared the performance of our model with a few state of the art object detectors (SNIPER\cite{singh2018sniper} and Yolo V3\cite{redmon2018yolov3}) on the compressed frames which is listed in Table \ref{tab:sota}. It is observed that our model displays relatively superior results.

We also observed that the model was capable of detection with different combinations of background and foreground movements Figure [\ref{fig:cars} and \ref{fig:persons}]. Our model could detect overlapping moving cars, just like other object-detection models.

Baraunink \etal \cite{Baraniuk2006} explains the connection between compressive sensing and Johnson-Lindenstrauss lemma. The preservation of distances between the higher-dimensional space to the lower-dimensional spaces can be the reason object detection in CS frames is possible using convolutional neural networks. The pre-trained weights of YOLO have equipped the model with the necessary features representations to differentiate different object classes.
We have also tested the model on coloured CS frames (Figure~\ref{fig:colour}). When trained on colour CS images it might work even better because of the extra features a model can learn. As colour CS CMOS cameras are not a distant reality, our results show that our model can work for them if needed in the future.

To evaluate the model, we deployed the network on NVIDIA TX2 on a typical surveillance use case. Here the the live camera feed from the camera was compressed using the same technique of pixel wise coded exposure mentioned as proposed in the dataset \cite{narayanan2019compressive}. These compressed frames were fed to the NVIDIA board where the model was deployed, the model took in the compressed frames as input and was able the predict the object which was present in them Figure \ref{fig:pipeline} depicts the process, the model provides the class of the object present with the location of object by surrounding them with a boundary box. We can observe that the compressed frame contains hazy information due to which the privacy concerns raising due to surveillance can be avoided. The compressed frame intactly preserves all the information required for the reconstruction of the frames but at the same time protects the privacy.

\section{Conclusion}
\label{sec:conclusion}

In this work, we showed that object detection and localization is possible directly at the compressed frame level in pixel-wise coded images. To our knowledge this is the first time the problem of object detection and localization in the CS domain has been attempted. Our model was also able to detect objects that could not be detected by humans in the CS domain. We also showed that our model works on different compression rates of CS frames. We reduced the time of object searching algorithms in surveillance videos by an order of 20x. As the CS cameras have low readout, real-time object detection is possible with our model. In future, we will continue adding more classes to the dataset and try other architectures to improve the precision of the model. We are looking at using the information of the sensing matrix to refine the results of the model. We envisage that this will be the beginning of object detection and other computer vision tasks in CS domain, which would make significant improvement in surveillance. Due to lack of any better comparison, Table \ref{tab:coco} shows the comparison of mAP values of existing object detection models in comparison to our model. The mAP of our proposed model suggest that its performing in par with that of the generic models that work on original domain, even though the boundaries and texture of the objects are not so well defined in the CS domain. The proposed model was able to detect and localize all the classes it has been trained on upon the compressed frame.

\begin{table}[!ht]
\begin{center}
\begin{tabular}{|l|c|c|c|}
\hline
Model & Dataset & mAP & FPS \\
\hline\hline
SSD300 & COCO & 41.2 & 46\\
SSD500 & COCO & 46.5 & 19 \\
YOLOv2 608x608 & COCO & 48.1 &40\\
YOLOv3 608 & COCO & 57.9 &51\\
Tiny YOLO & COCO & 23.7 & 244 \\
DSSD321 & COCO & 46.12 &12 \\
\hline
Proposed Model & Custom & 46.27 & 40 \\
\hline\hline

\end{tabular}
\end{center}
\caption{Mean Average Precision(mAP) of different models in comparison with custom and COCO dataset}
\label{tab:coco}
\end{table}





\bibliographystyle{unsrt}  


\end{document}